\documentclass{article}

\usepackage{arxiv}

\usepackage{xcolor}

\usepackage{amsthm}

\usepackage{algorithmic}

\newtheoremstyle{named}{}{}{\itshape}{}{\bfseries}{.}{.5em}{#1 \thmnote{#3}}
\theoremstyle{named}

\usepackage[utf8]{inputenc} 
\usepackage[T1]{fontenc}    
\usepackage{hyperref}       
\usepackage{url}            
\usepackage{booktabs}       
\usepackage{amsmath,amssymb,amsfonts}       
\usepackage{nicefrac}       
\usepackage{microtype}      
\usepackage{lipsum,cite}
\usepackage{graphicx,xcolor,amsmath,amssymb}
\usepackage{tikz}
\usepackage{siunitx}
\usepackage{ulem}
\usepackage{subcaption}
\usepackage{pgf-pie}
\usepackage[table]{xcolor}

\graphicspath{{Figures/}}
\usepackage{float}

\title{Error correction in multiclass image classification of facial emotion on unbalanced samples}

\iftrue

\author{
 Andrey A. Lebedev\\
  Lobachevsky State University of Nizhny Novgorod\\
  \texttt{stasenko@neuro.nnov.ru} 
 \And
  Victor B. Kazantsev\\
  Lobachevsky State University of Nizhny Novgorod\\
  Moscow Center for Advanced Studies\\
  \texttt{kazantsev@neuro.nnov.ru} \\
   \And   
 Sergey V. Stasenko\\
  Lobachevsky State University of Nizhny Novgorod\\
  Moscow Center for Advanced Studies\\
  \texttt{stasenko@neuro.nnov.ru} 
}

\fi

\begin{document}
\maketitle

\begin{abstract}
This paper considers the problem of error correction in multi-class classification of face images on unbalanced samples. The study is based on the analysis of a data frame containing images labeled by seven different emotional states of people of different ages. Particular attention is paid to the problem of class imbalance, in which some emotions significantly prevail over others. To solve the classification problem, a neural network model based on LSTM with an attention mechanism focusing on key areas of the face that are informative for emotion recognition is used. As part of the experiments, the model is trained on all possible configurations of subsets of six classes with subsequent error correction for the seventh class, excluded at the training stage. The results show that correction is possible for all classes, although the degree of success varies: some classes are better restored, others are worse. In addition, on the test sample, when correcting some classes, an increase in key quality metrics for small classes was recorded, which indicates the promise of the proposed approach in solving applied problems related to the search for rare events, for example, in anti-fraud systems. Thus, the proposed method can be effectively applied in facial expression analysis systems and in tasks requiring stable classification under skewed class distribution.
\end{abstract}


\section{Introduction}

Accurate facial emotion recognition has become a cornerstone in the broader domain of affective computing, which seeks to develop systems capable of recognizing, interpreting, and responding to human emotions using computational methods \cite{tao2005affective,calvo2010affect,liu2021emotion}. However, one persistent challenge in facial expression classification is the presence of imbalanced class distributions, where certain emotions are significantly underrepresented \cite{koelstra2012deap,sariyanidi2015automatic}. Class imbalance often degrades model generalization, leading to biased predictions toward majority classes and poorer recognition of subtle or rare emotions such as fear or disgust \cite{goodfellow2013challenges}.

To address such imbalances, researchers have proposed several strategies. Resampling techniques—such as oversampling minority classes or undersampling majority classes—are commonly employed to rebalance datasets, though they risk introducing redundancy or discarding critical information \cite{he2009learning,chawla2002smote}. Another avenue involves loss reweighting, where algorithms adjust the learning focus by assigning greater weight to underrepresented classes. Methods such as class-balanced loss, focal loss, and label-distribution-aware margins exemplify this approach \cite{cui2019class,lin2017focal,khan2019striking}. Hybrid approaches that combine reweighting with resampling have also been shown to improve robustness in imbalanced emotion datasets \cite{huang2019deep}.

Deep neural networks enhanced with attention mechanisms have demonstrated considerable promise in emotion recognition tasks. For instance, models integrating attention within CNN–LSTM architectures allow the system to focus on discriminative facial regions, improving performance \cite{hans2021cnn,rajpoot2022subject,wang2020region}. Furthermore, attention-based frameworks have been used explicitly to handle uncertainties and class imbalances by emphasizing learning from underrepresented samples \cite{saurav2025integrated,altalhan2025imbalanced}. Self-attention and Transformer-based designs have also gained traction for emotion recognition, outperforming traditional CNN-based approaches in scenarios with noisy or imbalanced data \cite{huang2021facial,daihong2021facial}.

Alongside these approaches, a complementary line of research emphasizes AI error correction without retraining legacy systems. This paradigm, pioneered by Gorban and colleagues, introduces external correctors—lightweight modules that detect risky cases and propose alternative decisions. The theoretical foundation rests on stochastic separation theorems, which demonstrate that in high-dimensional spaces, even simple classifiers (such as Fisher’s linear discriminants) can separate error cases from correctly classified ones with high probability \cite{gorban2023stochastic}. This leads to practical one-shot correction methods capable of improving system reliability without iterative retraining \cite{gorban2018correction}. More recent advances generalize these results to fine-grained, clustered data distributions, showing that families of multi-correctors can handle complex structures and even facilitate new class discovery in high-dimensional settings \cite{gorban2021entropy,grechuk2021general}. These methods have been validated on benchmarks such as CIFAR-10, where external correctors successfully amended errors of deep convolutional networks and supported incremental learning of novel object classes.

Despite these advances, most studies focus on models trained directly on the full emotion set, with balanced or rebalanced inputs, rather than exploring error correction strategies that recover missing or excluded classes. Your novel approach addresses this gap by using an LSTM-based model with attention that is trained on subsets of classes, with subsequent error correction applied to the omitted (seventh) class.

In your experiments, the model is trained on all possible combinations of six out of seven emotion classes, and then evaluated on its ability to correct for the seventh class that was held out during training. Your results show that while the success rate of correction varies per class, even small or rare emotion classes benefit, with key quality metrics—such as precision or recall—improving for those minority classes. These findings complement recent evidence showing that auxiliary learning and reconstruction-based strategies can significantly enhance minority class recognition in affective computing tasks \cite{yi2018data,chen2023semi}.

This finding is particularly significant for real-world applications where detecting rare or atypical emotional signals can be critical—for example, in anti-fraud systems or security contexts. The potential to enhance recognition of these low-frequency classes underscores the practical value of your method in scenarios requiring robust classification under skewed class distributions.
\section{Methodology}
\subsection{The corrector model}

Let's consider the problem of multiclass classification in the feature space. Let 
\[
\mathcal{X} \subseteq \mathbb{R}^n, 
\qquad 
\mathcal{Y} = \{1,2,\dots,K\}.
\]
The basic classifier is defined by the mapping
\[
f : \mathcal{X}\to \mathcal{Y}, 
\qquad 
x \mapsto f(x),
\]
which assigns a class label $f(x) to each point x\in\mathcal{X}\in\mathcal{Y}$.

In this case, the model additionally induces the display
\[
\mathbf{e} : \mathcal{X} \to \mathbb{R}^m,
\qquad 
x \mapsto \mathbf{e}(x),
\]
where $\mathbf{e}(x)$ is interpreted as an embedding vector obtained during the inference process.

To build a correction mechanism, we introduce the binary mapping
\[
g :\mathbb{R}^m\to\{0,1\},
\qquad 
\mathbf{z} \mapsto g(\mathbf{z}),
\]
where $g(\mathbf{z})=0$ corresponds to \emph{familiar} data, and $g(\mathbf{z})=1$ corresponds to \emph{unfamiliar} (out-of-distribution).

Based on this, we define a generalized classifier with correction:
\[
h :\mathcal{X}\to\mathcal{Y}\cup\{\text{new class}\},
\qquad
h(x) =
\begin{cases}
	f(x), & \text{if } g(\mathbf{e}(x)) = 0, \\[6pt]
	\text{new class}, & \text{if } g(\mathbf{e}(x)) = 1.
\end{cases}
\]

Thus, the $h$ function combines two mechanisms: 
\begin{enumerate}
	\item standard classification by means of $f$ on a set of known classes $\mathcal{Y}$;
	\item identification of new or missing classes using the $g$ corrector.
\end{enumerate}

This separation makes it possible to formalize the error correction task of a pre-trained system as a combination of the classification task and the task of detecting unknown data. In particular, the proposed approach provides:
\begin{itemize}
	\item resilience to the appearance of classes that were missing at the learning stage;
	\item the ability to process unbalanced samples by highlighting "rare" or excluded classes;
	\item extending the functionality of the original $f$classifier to the more general $h$ system.
\end{itemize}

\subsection{Metrics for evaluating the quality of a proofreader}

Let's give the test set $\mathcal{D}=\{(x_j,y_j)\}_{j=1}^N$, where $y_j\in\mathcal{Y}=\{1,\dots,K\}$.
Let's denote the predictions of the basic (without correction) model $f$ by $\hat y^{(f)}_j$,
and the predictions of the system with correction $h$ by $\hat y^{(h)}_j$.

For a fixed class $i\in\mathcal{Y}$, we assume
\[
J_i=\{j\in\{1,\dots,N\}: y_j=i\},\qquad N_i=|J_i|.
\]

Let's introduce a set of indexes of class i samples that have been correctly classified by one system or another: 
\[
A_i=\{j\in J_i:\ \hat y^{(f)}_j=i\}\quad\text{(correct by }f\text{)},
\]
\[
B_i=\{j\in J_i:\ \hat y^{(h)}_j=i\}\quad\text{(correct by }h\text{)}.
\]
We also denote the error sets (for class $i$):
\[
\bar A_i = J_i\setminus A_i\quad\text{(errors }f\text{on class }i\text{)},\qquad
\bar B_i = J_i\setminus B_i\quad\text{(errors }h\text{on class }i\text{)}.
\]

\subsubsection{Basic one-class metrics.}
First-class accuracy (synonyms of TPR for each class):
\[
\mathrm{TPR}^{(f)}_i=\frac{|A_i|}{N_i},\qquad
\mathrm{TPR}^{(h)}_i=\frac{|B_i|}{N_i}.
\]
Relative and absolute increment:
\[
\Delta_i=\mathrm{TPR}^{(h)}_i-\mathrm{TPR}^{(f)}_i,\qquad
R_i=\frac{\mathrm{TPR}^{(h)}_i}{\mathrm{TPR}^{(f)}_i+\varepsilon},
\]
where $\varepsilon>0$ is a small regularization for stability at zero base.

\paragraph{Conservation and harm metrics for already correctly classified samples.}
The characteristic we are interested in is how the corrector affects those samples that have already been correctly classified by $f$. We introduce the following concepts.

\textbf{Retention} — the proportion of samples of class $i$ that were correctly recognized by $f$, which remained correctly recognized even after applying the corrector:
\[
\mathrm{Ret}_i \;=\; \frac{|A_i\cap B_i|}{|A_i|}\quad\text{(if }|A_i|>0\text{)}.
\]

\textbf{Harm (harm)} — the proportion of $i$ class samples correctly recognized by $f$ that became erroneous after correction:
\[
\mathrm{Harm}_i \;=\; \frac{|A_i\setminus B_i|}{|A_i|} \;=\; 1-\mathrm{Ret}_i.
\]

\textbf{Correction gain} — the proportion of previously misclassified Class i samples that became correct after correction:
\[
\mathrm{Gain}_i \;=\; \frac{|B_i\setminus A_i|}{|\bar A_i|}\quad\text{(if }|\bar A_i|>0\text{)}.
\]


\subsubsection{Metrics of "spillover" and false positives on other classes.}
To take into account how the corrector affects the incorrect reassignment of the label towards the class $i$, we define
\[
\mathrm{Spill}_i \;=\; \frac{|\{j:\ y_j\neq i,\ \hat y^{(h)}_j = i\}|}{N-N_i}
\]
— the proportion of samples from other classes that were classified as class i by the $h$ system (i.e., "false positive for i"). Accordingly, for $f$:
\[
\mathrm{FPR}^{(f)}_i \;=\; \frac{|\{j:\ y_j\neq i,\ \hat y^{(f)}_j = i\}|}{N-N_i},
\qquad
\mathrm{FPR}^{(h)}_i \;=\; \frac{|\{j:\ y_j\neq i,\ \hat y^{(h)}_j = i\}|}{N-N_i}.
\]
False-positive percentage change:
\[
\Delta\mathrm{FPR}_i \;=\; \mathrm{FPR}^{(h)}_i - \mathrm{FPR}^{(f)}_i.
\]

\subsubsection{Summary indicators and standards.}
The following aggregates are offered for a summary assessment:
\[
\overline{\mathrm{Ret}}=\frac{1}{K}\sum_{i=1}^K \mathrm{Ret}_i,\qquad
\overline{\mathrm{Harm}}=\frac{1}{K}\sum_{i=1}^K \mathrm{Harm}_i,
\]
\[
\overline{\mathrm{Gain}}=\frac{1}{K}\sum_{i=1}^K \mathrm{Gain}_i,\qquad
\overline{\Delta\mathrm{FPR}}=\frac{1}{K}\sum_{i=1}^K \Delta\mathrm{FPR}_i.
\]
You can also weight the frequencies of the classes $N_i/N$, if desired, to get a "micro" version of the metrics:
\[
\mathrm{Ret}^{\mathrm{(w)}}=\sum_{i=1}^K\frac{N_i}{N}\mathrm{Ret}_i,\quad\text{etc.}
\]

\subsubsection{Interpretation of metrics.}
\begin{itemize}
	\item $\mathrm{Ret}_i$ close to 1 means that the corrector does not spoil the already correctly recognized samples of class $i$.
	\item $\mathrm{Harm}_i$ demonstrates the share of "collateral damage" for already correct predictions; the lower the better.
	\item $\mathrm{Gain}_i$ shows the corrector's ability to restore previously erroneous examples of this class.
	\item$\Delta\mathrm{FPR}_i>0$ indicates an increase in false-positive assignments to class $i$ (side effect).
\end{itemize}

\subsection*{Statement of the problem}

In this paper, we propose a method for error correction in the problem of multi-class classification of facial images with unbalanced samples using an ensemble of a neural network model and a gradient boosting algorithm. The experiments were performed on a dataset containing data on 7 classes of emotional states of people of different ages with a pronounced imbalance in the distribution by classes. The method includes the following stages:

\textbf{Iterative exclusion of classes.} For each experiment, one class is excluded from the training set. The model is trained on the remaining six classes, and the excluded one is used at the inference and correction stage. Thus, a sequential analysis of the model's behavior is carried out in the absence of each of the classes.

\textbf{Base model: LSTM with an attention mechanism.} To process images, a model with an LSTM architecture and an attention mechanism is used, capable of focusing on key areas of the face. Images are pre-encoded by convolutional layers, after which they are passed to the LSTM model.

\textbf{Feature extraction from latent variables.} After the neural network inference, all latent variables are saved — including the outputs of convolutional layers, attention vectors, LSTM hidden states, and other intermediate representations. This allows us to obtain a detailed description of the model's response to each input stimulus.

\textbf{Training the corrector model (gradient boosting).} At the correction stage, the gradient boosting algorithm is used, trained on the full set of extracted features. The goal is to correct the neural network errors on previously unseen examples from the excluded class. Boosting is trained on the neural network inferences and is able to identify patterns that were not taken into account by the first model due to their absence.

\textbf{Accuracy analysis by classes.} After the correction, the accuracy of predictions is estimated for each class separately, including both those presented in training and the previously excluded one. This allows us to determine how effectively each specific class is being restored and to identify the best configurations for correction.

\subsection*{Data}
For analysis, the RAF-DB (Real-world Affective Faces Database) dataset was selected, containing data on the emotions of people of different ages \cite{li2017reliable}.

\begin{figure}[h!]
	\centering
	\includegraphics[width=0.95\textwidth]{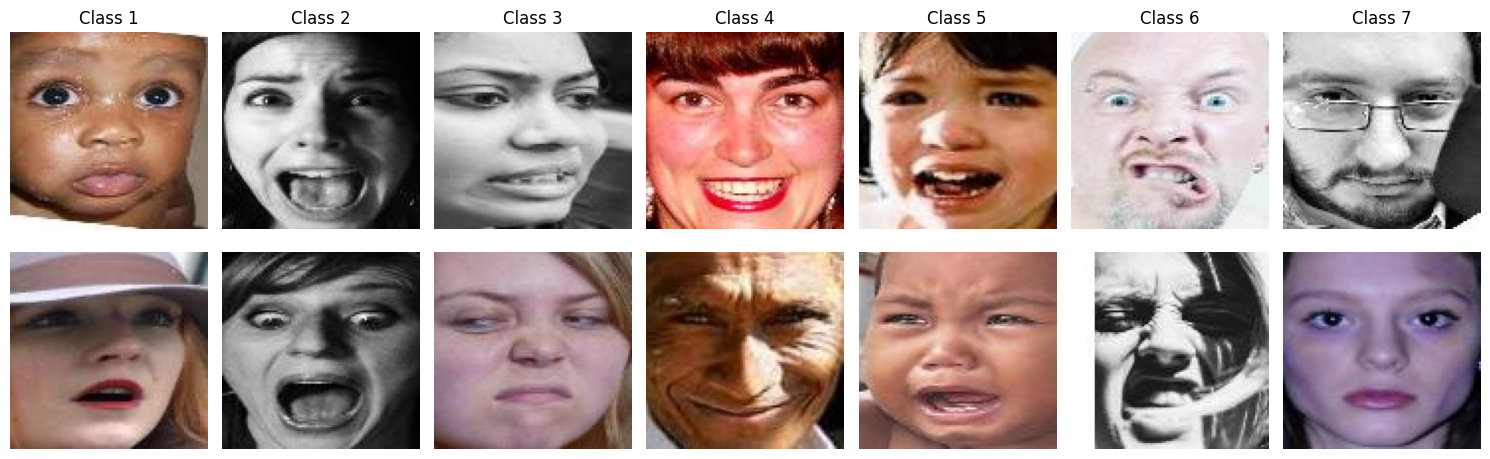}
	\caption{Examples of images labeled with different emotions in the sample, according to the labeling shown in Table ~\ref{tab:data_distribution}. 
	}
	\label{pic:emotion_examples}
\end{figure}

Examples of images are shown in Fig ~\ref{pic:emotion_examples}.
The dataset contains 15,339 facial images labeled with basic and composite emotions by 40 independent annotators. The label mapper, which shows the face label and the emotion it encodes, is shown in ~\ref{tab:data_distribution}. The images are highly diverse in features such as age, gender, ethnicity, head rotation, lighting conditions, and the presence of partial occlusions (e.g. glasses, beard, or self-occlusion). In addition, the images may contain filters and special effects, which makes the dataset particularly suitable for emotion recognition tasks in near-real-world settings.

\begin{table}
	\centering
	\begin{tabular}{|c|c|c|c|c|c|}
		\hline
		& & \multicolumn{3}{c}{\textbf{Data Split}} & \\
		\hline
		\textbf{Label} & \textbf{Emotion} & \textbf{Train} & \textbf{Test} & \textbf{Correct} & \textbf{Sum} \\
		\hline
		1 & Surprise & 814 & 379 & 426 & \textbf{1620} \\
		\hline
		2 & Fear & 166 & 88 & 101 & \textbf{357} \\
		\hline
		3 & Disgust & 449 & 220 & 208 & \textbf{880} \\
		\hline
		4 & Happiness & 2974 & 1506 & 1477 & \textbf{5961} \\
		\hline
		5 & Sadness & 1216 & 628 & 616 & \textbf{2465} \\
		\hline
		6 & Anger & 435 & 216 & 216 & \textbf{873} \\
		\hline
		7 & Neutral & 1615 & 798 & 791 & \textbf{3211} \\
		\hline
		\multicolumn{2}{|c|}{\textbf{All emotions}} & \textbf{7669} & \textbf{3835} & \textbf{3835} & \textbf{15339} \\
		\hline
	\end{tabular}
	\caption{Correspondence between labels and emotions in the RAF-DB dataset (basic emotions)}
	\label{tab:data_distribution}
\end{table}

To conduct experiments according to the described methodology, it is necessary to select three random subsets from the initial data:

\begin{itemize}
	\item \textbf{Train} — data for training the LSTM model regardless of the class configuration;
	\item \textbf{Correct} — data for training the corrector model;
	\item \textbf{Test/Val} — data for testing and assessing the accuracy of the models.
\end{itemize}

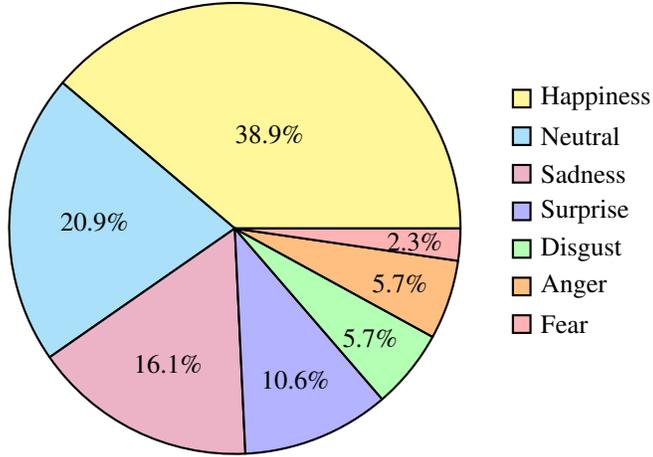
\begin{figure}[h]
	\centering
\begin{tikzpicture}
	\pie[
	radius=3,
	sum=auto,
	after number=\%, 
	text=legend, 
	color={yellow!50, cyan!30, purple!30, blue!30, green!30, orange!50, red!30}
	]
	{38.9/Happiness, 20.9/Neutral, 16.1/Sadness, 10.6/Surprise, 5.7/Disgust, 5.7/Anger, 2.3/Fear}
\end{tikzpicture}
	\caption{Pie chart showing the proportion of classes in the initial sample.}
	\label{fig:data_distribution}
\end{figure}

Distributions of the proportions of classes in each subset are given in figure~\ref{fig:data_distribution}.

\section{Technical implementation of the experiment}

To solve the problem of classifying emotions in facial images, a custom neural network model was developed that combines convolutional layers for feature extraction, a recurrent component (LSTM) for modeling the spatial-sequential structure, and an attention mechanism for identifying the most informative features.

\begin{figure}[h!]
	\centering
	\includegraphics[width=1\textwidth]{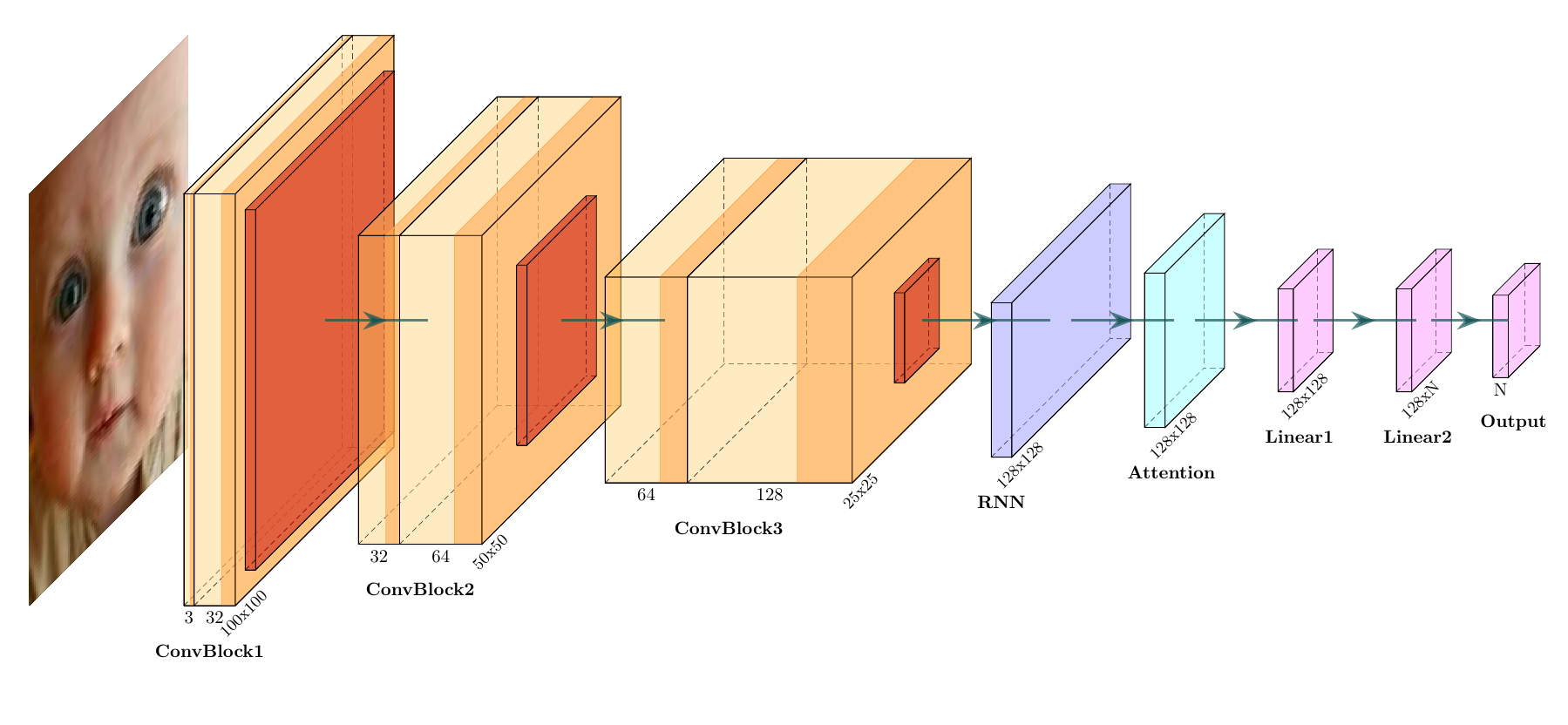}
	\caption{Scheme of the developed convolutional neural network model with memory and attention }
	\label{fig:model_scheme}
\end{figure}

The model consists of the following main components:
\subsection{Feature extraction block / Convolutional block (\texttt{conv\_layers})}

Consists of three convolutional blocks, each of which includes:
\begin{itemize}
	\item convolutional layer (\texttt{Conv2d}) with kernel \( 3 \times 3 \), stride 1 and padding 1;
	\item batch normalization (\texttt{BatchNorm2d});
	\item ReLU activation function;
	\item pooling layer \texttt{MaxPooling} with kernel \( 2 \times 2 \).
\end{itemize}

Layer parameters:
\begin{itemize}
	\item Conv1: \texttt{in\_channels=3}, \texttt{out\_channels=32};
	\item Conv2: \texttt{in\_channels=32}, \texttt{out\_channels=64};
	\item Conv3: \texttt{in\_channels=64}, \texttt{out\_channels=128}.
\end{itemize}

At the output of the block, spatial feature maps of dimension \( [B, 128, H', W'] \), where \( B \) is the batch size, are formed.

\subsection{NN block}
\subsubsection{Recurrent Block (\texttt{rnn})}

The output of the convolutional block is transformed into a sequence: the tensor is transformed into the form \( [B, T, D] \), where \( T = H' \cdot W' \), and \( D = 128 \). On this sequence, an LSTM is applied with the following parameters:

\begin{itemize}
	\item input and hidden state size: 128;
	\item \texttt{batch\_first=True}.
\end{itemize}

This allows the model to capture spatial dependencies between different regions of the image.

\subsubsection{Attention Mechanism (\texttt{attention})}

The output of the LSTM is a multi-headed attention mechanism (\texttt{MultiheadAttention}):

\begin{itemize}
	\item key, query, and value dimensions: 128;
	\item one projection layer (\texttt{out\_proj}) is used at the output.
\end{itemize}

The attention mechanism allows the model to focus on the most significant areas of the face when making a classification decision.

\subsection{Classification head / Fully connected classifier (\texttt{fc\_layers})}

The attention result is aggregated and fed to the fully connected part of the model:

\begin{itemize}
	\item \texttt{Linear(128 $\rightarrow$ 128)} with ReLU activation;
	\item \texttt{Dropout(p=0.5)} for regularization;
	\item \texttt{Linear(128 $\rightarrow$ 7)} is the output layer for classification by N emotions.
\end{itemize}

\subsection{Loss function}

Since there is a pronounced imbalance between the classes in the training set, a weighted loss function is used. The class weights are calculated as the ratio of the total number of examples to the number of examples in each class:
\[
w_i = \frac{N}{n_i},
\]
where \( N \) is the total number of samples, \( n_i \) is the number of samples of the \( i \)-th class.

Next, the cross-entropy loss function is used:
\[
\mathcal{L} = - \sum_{i=1}^{C} w_i \cdot y_i \log(\hat{y}_i),
\]
where \( C \) is the number of classes, \( y_i \) is the true label, 
\( \hat{y}_i \) is the model output (softmax probability) for class \( i \), 
and \( w_i \) is the class weight.

\subsection{Early stopping mechanism}
To determine the criterion for early learning stoppage, we tried several different rules based on the average roc-auc-score, loss function, and average accuracy. It has been empirically found that the best rule for stopping learning in the proposed architecture is:
\begin{itemize}
	\item loss function on train set $\rightarrow$ min;
	\item accuracy on the validation set $\rightarrow$ max.
\end{itemize}

\subsection{Model training.}
The results of a study of a descriptive neural network in conjugation of both grades 6 and 7 showed, in the form of a segmented metric, the accuracy of determining classes in the table~\ref{tab:full_correct_report}.

Accuracy is calculated as the proportion of correctly predicted labels:
\[
\text{Accuracy} = \frac{1}{N} \sum_{i=1}^{N} \mathbb{I}(y_i = \hat{y}_i),
\]
where \( \mathbb{I} \) is the indicator function.

\subsection{Extraction of hidden features}

For correction, a method of repeated inference of the model on additional images 
that were not included in the training and validation samples was used. 
During the inference process, the values of all intermediate representations 
(feature maps and hidden states) were saved at the stages:
\begin{itemize}
	\item output from convolutional layers;
	\item output from the LSTM layer;
	\item after applying the attention mechanism (Multihead Attention);
	\item output from the fully connected layer before the final classification.
\end{itemize}

Thus, for each image, a vector of hidden features was formed, containing multi-level information about its spatial and semantic content.

\section{Mechanism for Correcting Neural Network Predictions}

After completing the training of the basic neural network model, an additional stage of analysis and correction of predictions was implemented in order to improve the quality of classification under uncertainty on new data.

\subsection{Training a Gradient Boosted Model}

The generated latent feature vectors were used as input for a gradient boosting model (e.g., \texttt{XGBoost} or \texttt{LightGBM}). The target variable was the true class label of the image.

A special feature of the approach is that the boosting model is trained on examples that were not used to train the original neural network. This allows boosting to identify patterns in the latent representations that were not covered by LSTM training, and thus compensate for possible gaps in the generalization ability of the base model.

\subsection{Training the corrector.}
To train the corrector, we used the XGBoost implementation with parameters selected empirically.

\subsection{Using the Corrector}

After training, the corrector model was used as an additional filter: during inference, the main model passes the hidden features to boosting, and, in case of a high error probability ($g(\mathbf{z}_i) < \tau$), the prediction is either corrected or not depending on the decision policy.

\subsection{Purpose of the method}

Thus, gradient boosting acts as an external evaluator over the outputs of the neural network. It can:

\begin{itemize}
	\item improve the accuracy of class recognition on previously unseen data;
	\item identify areas of low confidence in the neural network;
	\item use more flexible dependencies between latent features and class labels.
\end{itemize}

This approach can be viewed as a type of stacking, where boosting serves as a meta-model trained on the embeddings of the base neural network.

\section{Results}

This section presents the quantitative results of applying the proposed classification error correction method. 
The analysis was carried out using the metrics \textit{Retention}, \textit{Harm}, changes in \textit{Gain}, \(\Delta\mathrm{FPR}\),
as well as the integral performance characteristics listed below.
\begin{table}[h!]
	\centering
	\rowcolors{2}{white}{gray!10}
	\begin{tabular}{|c|*{7}{c}|}
		\hline
		\multicolumn{8}{|c|}{\cellcolor{green!15}\textbf{Retention}} \\
		\hline
		& \multicolumn{7}{c|}{Corrector} \\
		\hline
		\textbf{Label} & 1 & 2 & 3 & 4 & 5 & 6 & 7 \\
		\hline
		1 & \underline{1.000} & 1.000 & 1.000 & 0.963 & 0.995 & 0.997 & 0.971 \\
		2 & \cellcolor{red!20}0.898 & \underline{1.000} & 0.977 & 0.989 & 1.000 & \cellcolor{red!20}0.977 & 1.000 \\
		3 & 1.000 & 1.000 & \underline{1.000} & 0.977 & 1.000 & 1.000 & \cellcolor{red!20}0.945 \\
		4 & 0.999 & 0.999 & 0.999 & \underline{1.000} & 0.997 & 0.999 & 0.975 \\
		5 & 0.992 & 1.000 & \cellcolor{red!20}0.995 & \cellcolor{red!20}0.876 & \underline{1.000} & 1.000 & \cellcolor{red!20}0.881 \\
		6 & 0.981 & \cellcolor{red!20}0.995 & 1.000 & \cellcolor{red!20}0.903 & \cellcolor{red!20}0.968 & \underline{1.000} & 0.977 \\
		7 & 0.982 & 0.999 & 0.997 & 0.955 & \cellcolor{red!20}0.957 & 1.000 & \underline{1.000} \\
		Average & 0.973 & 0.999 & 0.996 & 0.940 & 0.995 & 0.996 & 0.963 \\
		\hline
	\end{tabular}
	\begin{tabular}{|c|*{7}{c}|}
		\hline
		\multicolumn{8}{|c|}{\cellcolor{red!15}\textbf{Harm}} \\
		\hline
		& \multicolumn{7}{c|}{Corrector} \\
		\hline
		\textbf{Label} & 1 & 2 & 3 & 4 & 5 & 6 & 7 \\
		\hline
		1 & \underline{0.000} & 0.000 & 0.000 & 0.037 & 0.005 & 0.003 & 0.029 \\
		2 & \cellcolor{red!20}0.102 & \underline{0.000} & 0.023 & 0.011 & 0.000 & \cellcolor{red!20}0.023 & 0.000 \\
		3 & 0.000 & 0.000 & \underline{0.000} & 0.023 & 0.000 & 0.000 & \cellcolor{red!20}0.055 \\
		4 & 0.001 & 0.001 & 0.001 & \underline{0.000} & 0.003 & 0.001 & 0.025 \\
		5 & 0.008 & 0.000 & \cellcolor{red!20}0.005 & \cellcolor{red!20}0.124 & \underline{0.000} & 0.000 & \cellcolor{red!20}0.119 \\
		6 & 0.019 & \cellcolor{red!20}0.005 & 0.000 & \cellcolor{red!20}0.097 & \cellcolor{red!20}0.032 & \underline{0.000} & 0.023 \\
		7 & 0.018 & 0.001 & 0.003 & 0.045 & \cellcolor{red!20}0.043 & 0.000 & \underline{0.000} \\
		\rowcolor{gray!10}
		Average & 0.021 & 0.001 & 0.005 & 0.048 & 0.012 & 0.004 & 0.036 \\
		\hline
	\end{tabular}
	
	\caption{Values of \textit{Harm} and \textit{Retention} metrics for all classes. 
		The Harm metric reflects the proportion of examples that were correctly classified by the original model,
		but became erroneous after applying the corrector. 
		The Retention metric shows the proportion of correct predictions that remained after the correction. 
		The values on the main diagonal are underlined and correspond to preserving/distorting predictions within the same class. The Retention and Harm metrics are also highlighted in red, which are higher and lower than the average, respectively.}
	\label{tab:harm_retention}
\end{table}

\begin{table}[h!]
	\centering
	\rowcolors{2}{white}{gray!10}
	\begin{tabular}{ccc}
		\toprule
		\textbf{Label} & $\overline{\Delta\mathrm{FPR}}$ & $\overline{\mathrm{Gain}}$ \\ 
		\midrule
		1 & 0.015 & 0.517 \\ 
		2 & 0.001 & 0.239 \\ 
		3 & 0.029 & 0.482 \\ 
		4 & 0.095 & 0.811 \\ 
		5 & 0.023 & 0.242 \\ 
		6 & 0.002 & 0.204 \\ 
		7 & 0.072 & 0.590 \\ 
		\bottomrule
	\end{tabular}
	\caption{Summary metrics of proofreader quality for each class. 
		\(\overline{\Delta\mathrm{FPR}}\) — the average change in the false positive error after applying the corrector. 
		\(\overline{\mathrm{Gain}}\) — the proportion of previously erroneous examples of the corrected class that were classified correctly.}
	\label{tab:gain_net}
\end{table}

From the results shown in the \ref{tab:harm_retention} table, it can be seen that the values of \textit{Retention} remain high (close to unity) in all cases,
which indicates that most of the correct predictions of the basic model are preserved. 
The values of \textit{Harm}, on the contrary, record cases when the corrector introduces distortions. From the table \ref{tab:harm_retention} it can be seen that the corrector does the most harm to the recognition of Fear(2) when correcting Surprise(1), Sadness(5) and Anger(6) when correcting Happiness(4) and Sadness(5) when correcting Neutral(7)

Summary indicators of the corrector's effectiveness are given in the table~\ref{tab:gain_net}. 
There is a positive trend for all classes \textit{Gain}, which confirms the correctors ability
to improve the completeness of the classification of the excluded class. 
The growth of \(\Delta\mathrm{FPR}\) remains relatively moderate, which makes the increase in \textit{Gain} statistically significant.
The greatest increase in quality is observed for <<Happiness>>(4), <<Surprise>>(1), <<Disgusting>>(3), <<Neutral>>(7) and even worse <<Fear>>(2), <<Sadness>>(5), <<Anger>>(6).

The table~\ref{tab:full_correct_report} shows the results of experiments to exclude classes from the training sample. 
In different cases, the correction allows for varying degrees of partial compensation for the loss of information about the excluded class.

\begin{table}[h!]
	\centering
	\rowcolors{2}{white}{gray!10}
	\begin{tabular}{c ccccccc c c}
		\toprule
		\textbf{Predict} & \multicolumn{7}{c}{\textbf{Correct}} & \textbf{Not correct} & \textbf{P} \\
		\cmidrule(lr){2-8}
		\textbf{Label} & \textbf{1} & \textbf{2} & \textbf{3} & \textbf{4} & \textbf{5} & \textbf{6} & \textbf{7} & & \\
		\midrule
		1 & \cellcolor{red!20}0.51 & 0.77 & 0.79 & \underline{0.84} & 0.75 & 0.76 & 0.80 & 0.78 & \cellcolor{orange!20}0.55 \\
		2 & \underline{0.55} & \cellcolor{red!20}0.24 & 0.06 & 0.43 & 0.47 & 0.45 & 0.39 & 0.43 & \cellcolor{orange!20}0.56 \\
		3 & 0.38 & 0.50 & \cellcolor{red!20}0.48 & \underline{0.98} & 0.98 & 0.97 & 0.95 & 0.74 & \cellcolor{yellow!20}0.65 \\
		4 & 0.99 & \underline{0.99} & 0.99 & \cellcolor{red!20}0.77 & 0.86 & 0.90 & 0.92 & 0.90 & \cellcolor{green!20}0.86 \\
		5 & 0.69 & 0.71 & 0.72 & 0.71 & \cellcolor{red!20}0.24 & 0.72 & \underline{0.80} & 0.78 & \cellcolor{red!20}0.31 \\
		6 & 0.68 & 0.66 & 0.63 & 0.63 & \underline{0.73} & \cellcolor{red!20}0.20 & 0.63 & 0.67 & \cellcolor{red!20}0.30 \\
		7 & 0.77 & 0.76 & 0.73 & 0.79 & 0.87 & 0.71 & \cellcolor{red!20}0.59 & \underline{0.90} & \cellcolor{yellow!20}0.66 \\
		\bottomrule
	\end{tabular}
	
	\caption{Comparison of classification accuracy when using a corrector for different combinations of training classes. 
		Values on the main diagonal (errors within its own class) highlighted in red, 
		the maximum values by row (the best prediction option for this class) are green. 
		The last column shows the accuracy of the original model without correction.}
	\label{tab:full_correct_report}
\end{table}

For a formal assessment of the corrector's contribution to the final quality of the classification, we introduce 
corrector power indicator \(P\), which is defined as the ratio of classification accuracy with correction to accuracy without correction:
\[
P=\frac{\mathrm{Accuracy}_{\text{with corr}}}{\mathrm{Accuracy}_{\text{without corr}}}.
\]

The table~\ref{tab:full_correct_report} shows the values of the strength indicator of the corrector \(P\).
The best result was recorded for the class <<Happiness>> (\(P = 0.86\)),
while for the classes <<Sadness>> and <<Anger>> the values \(P\) is significantly lower (\(P\approx 0.30\)). 
This indicates a pronounced class dependence of the method's effectiveness:
for some emotions, correction successfully restores accuracy,
for others, the increase is lower.

Thus, the error correction method is a promising tool. It is capable of increasing the stability of multiclass classifiers in conditions of incomplete and unbalanced samples,
but its effectiveness depends on the class structure and requires further research.

\section{Discussion}
However, in the general case, correction turns out to be less effective than complete retraining and does not allow reaching the level of a fully trained model. In turn, building a corrector is computationally much easier than completely retraining the model, as well as the proposed method allows you to create many correctors that implement a complex, nonlinear stack model.
Further development of the approach may go in the direction of:
\begin{itemize}
	\item for using more compact embeddings;
	\item scaling experiments on large samples to increase stability;
	\item for developing adaptive correctors that take into account the semantic and statistical relationships of classes.
\end{itemize}


\section{Conclusion}
The results show that the proposed error correction method can significantly improve the quality
of recognition of excluded classes, while maintaining high accuracy on already known data. 
However, a significant dependence of efficiency on the nature of a particular class has been revealed.

High values of \textit{Retention} indicate that the basic structure of the model's predictions is preserved., 
and the risk of quality deterioration for the initial classes remains low. 
At the same time, the values \textit{Harm} and \(\Delta\mathrm{FPR}\) demonstrate, 
that the corrector can introduce undesirable distortions, especially in the case of emotions that are similar in feature space. 
Thus, the method has a positive effect, but requires careful adjustment.

Experiments on class exclusion show that the corrector works especially successfully for classes with pronounced interclass similarity (for example, <<Happiness>>), whereas for classes less related to others (for example, <<Sadness>> and <<Anger>>), the effectiveness is noticeably lower.

\section{Acknowledgements}

The work was carried out with the support of the federal assignment of the Ministry of Science and Higher Education of the Russian Federation, project No. FSMG-2024-0047.





\end{document}